\documentclass[letterpaper, 10 pt, conference]{ieeeconf}  

\IEEEoverridecommandlockouts                              

\usepackage{xcolor}
\usepackage{marginnote}
\usepackage{soul}
\usepackage{cite}

\title{\LARGE \bf
Optometrist's Algorithm for Personalizing Robot-Human Handovers}

\author{Vivek Gupte$^{1}$, Dan R. Suissa$^{2}$ and Yael Edan$^{3}$
\thanks{$^{1}$Vivek Gupte is with the Department of Mechanical Engineering, BITS-Pilani, K. K. Birla Goa Campus, India.
        {\tt\small f20180948@goa.bits-pilani.ac.in}}%
\thanks{$^{2}$Corresponding Author. Dan R. Suissa is with the Department of Computer Science, Ben-Gurion University of the Negev, Israel
        {\tt\small danrouve@bgu.ac.il}}%
\thanks{$^{3}$Yael Edan is with the Department of Industrial Engineering and Management, Ben-Gurion University of the Negev, Israel
        {\tt\small yael@bgu.ac.il}}%
}

\begin{document}

\maketitle

\thispagestyle{empty}
\pagestyle{empty}

\begin{abstract}

With an increasing interest in human-robot collaboration, there is a need to develop robot behavior while keeping the human user's preferences in mind. Highly skilled human users doing delicate tasks require their robot partners to behave according to their work habits and task constraints. To achieve this, we present the use of the Optometrist's Algorithm (OA) to interactively and intuitively personalize robot-human handovers. Using this algorithm, we tune controller parameters for speed, location, and effort. We study the differences in the fluency of the handovers before and after tuning and the subjective perception of this process in a study of $N=30$ non-expert users of mixed background -- evaluating the OA. The users evaluate the interaction on trust, safety, and workload scales, amongst other measures. They assess our tuning process to be engaging and easy to use. Personalization leads to an increase in the fluency of the interaction. Our participants utilize the wide range of parameters ending up with their unique personalized handover.
\end{abstract}

\section{INTRODUCTION}
There has been an increased interest in incorporating robots to accomplish specific tasks where human skill and experience are valuable. In such tasks, robots can work as assistants. Manipulators can provide the skilled user with an extra arm to hold more objects, move far-away objects closer and perform many other motions per the task requirements. While collaborating with a skilled human performing delicate tasks, it is preferred that incorporating robotic assistants does not require the human user to change the habits and techniques they have developed over the years. Instead, the robot may adapt its behavior to match the human partner.

Using Human-In-The-Loop control systems where the human can directly or indirectly influence the control signal is a promising approach to this problem. Such methods allow for a customized human-robot interaction. These controllers can be classical\cite{micelli} or predictive systems \cite{yang}.
However, expert knowledge and an understanding of the system's parameters are often required to tune a robot controller to one's liking \cite{simon}. Moreover, tuning of parameters is usually achieved by directly selecting parameter values through some interface\cite{alap-timingspec, alili}. Even with expert knowledge, directly choosing the preferred parameter values can be difficult and time-consuming\cite{simon}.

\begin{figure}[ht]
	\centering
        \vspace{4pt}
		\includegraphics[scale=0.066]{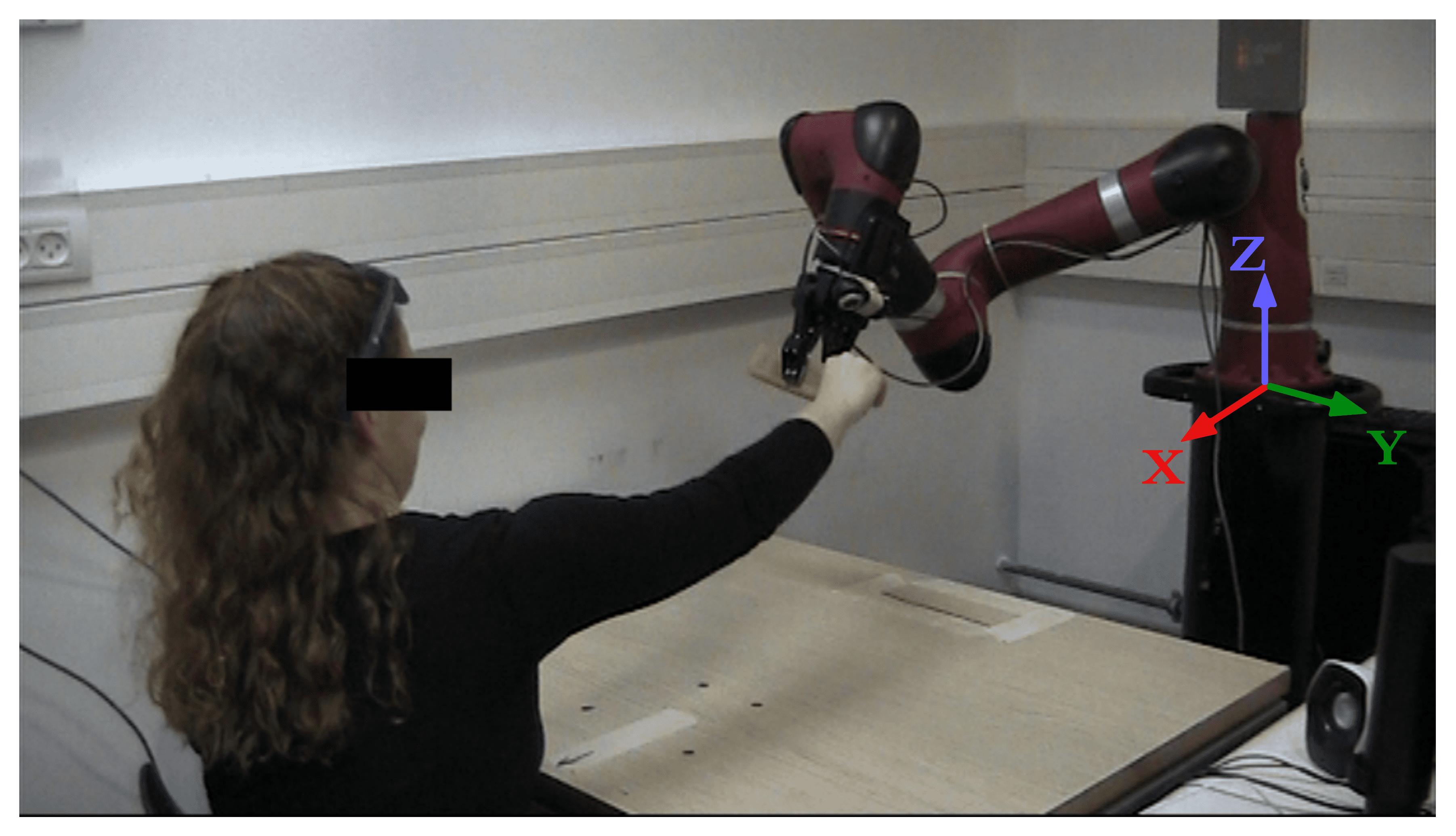}
	\caption[]{A robot-human handover between a user and the Sawyer Robot.}
	\label{fig:Robot-human handover}
    \vspace{-0.5cm}
\end{figure}

This work focuses on personalizing a controller for a collaborative task -- robot-human handover.
We introduce, implement and analyze a personalization algorithm -- \textit{The Optometrist's Algorithm} -- which allows non-expert users intuitive access to parameter tuning.
A robot-human handover is the transfer of an object from the robot to a human collaborating with the robot\cite{constanzo}.
We choose handovers as it is a widely studied fundamental robot skill with numerous applications in various situations.

\section{RELATED WORKS}
In the effort to make robot-human handovers more natural and fluent, different control systems, e.g., \cite{alap-syn, micelli, medina} and learning frameworks, e.g., \cite{minwu(dmp), alap-gps, kupicsk}, have been established. 

        A study by Cakmak et al.\cite{cakmak} shows the importance of involving humans in the personalizing processes. The authors report that their participants preferred handover configurations learned from human examples over planned configurations, despite the planned configurations being more efficient regarding objective metrics. 

A human-feedback controller for handovers was introduced by Kshirsagar et al.\cite{alap-timingspec}. In this work, the human user gave the robot feedback based on task constraints. The participants directly tuned the parameters of the robot controller by choosing parameter values that should be used in the particular task through a GUI (sliders). The participants of this study described the need for a more accessible way to tune robot parameters.

Kupcisk et al.\cite{kupicsk} have used human preferences and evaluative feedback to train a handover. They used the feedback to estimate a latent reward function and learned the handover using contextual policy search. This study tuned several non-intuitive control parameters, such as compliance and grip force. The feedback consisted of an absolute factor, which graded the interaction, and a preferential factor which comparatively rated the interaction against earlier ones. The authors have reported establishing that absolute feedback is used preferably when decisions are obvious (failure cases, very bad controller, excellent controller). In contrast, preference-based feedback is more accessible for human users to assess.

Building upon these insights, we present our method to tune robot-human handovers, according to user preferences, with little interactions\footnote{Depending on the problem size and search space} and no need to understand or directly choose any values for any of the parameters involved. We present an easy-to-use method -- the Optometrist's Algorithm (OA) -- which allows a non-expert user to indirectly tune several parameters of a handover controller. The OA accomplishes the tuning procedure inspired by the method when one chooses their glasses. We provide evidence that this type of comparison-based feedback allows for an intuitive tuning process, even when the parameters are hard to understand or when multiple parameters need to be tuned (sequentially\footnote{Note, the algorithm also allows for parallel tuning of multiple parameters, which we did not use in this study.}). 

While preparing the manuscript, we found the implementation of a similar algorithm for tuning experimental parameters in a Fusion experiment\cite{fusion}. Our work introduces a distilled version of the same concept into robotics and HRI for personalizing a robot controller. Specifically, we study the use of this algorithm in tuning a handover and its effects on objective and subjective task metrics such as success and fluency. Furthermore, we assess the personalization of our controller and if any parameters affect the user more than the others. We also investigate whether our participants can identify slight variations in their tuned handovers after working for a while with the robot to assess the personalization.

\section{METHOD}

In this study, we developed a customized robot-human handover controller for an object of fixed shape. The participants personalized this controller using the Optometrist's Algorithm (OA) by repeatedly choosing between two options (comparative feedback).

\subsection{Handover Controller}
The handover motion was developed using the Sawyer Robot's motion controller. The reach actions were based on 'ROS MoveIt!'. The robot end-effector was commanded to reach the necessary coordinates in Cartesian space. A handwritten script controlled the robot's actions and phases of the handover.

The robot picked up the object from a table (pick-up location) and moved to the starting location (both poses predefined and fixed throughout all experiments). From there, it began the 'reach' phase of the handover and moved towards the user (location of the handover - proximity, side, and height $[x,y,z]$ - as well as the speed of the reach $V_{max}$ was tunable through the OA). Once the robot stopped, it waited for the user to initiate the object transfer. The user could take the object from the robot's grasp by overcoming a force threshold ($F_{min}$, tunable through OA). Once the object was released, the robot moved back above the pick-up location. Meanwhile, the user had to restore the object to the pick-up location and prepare for the subsequent handover. We chose this method of determining a handover location according to a fixed pose defined by parameters over others (like computer vision), as it standardizes the handover and its tuning for each sample and each participant. Also, the need to tune a pose in the robot's reference frame provides us with a non-intuitive parameter.

The parameters that we tuned were as given in Table I. For each parameter, a lower and higher limit and a step size were determined empirically before the experiment, resulting in five sets of roughly ten parameter values each - to be tuned by the OA.

\begin{table}[h!]
\begin{center}
\begin{tabular}{|c|c|c|c|}
 \hline
 Parameter & Min & Step & Max\\ 
 \hline
 $V_{max} [m/s]$  & 0.1 & 0.1 & 0.8  \\  
 $x [m] $ & 0.8 & 0.025 & 1.0 \\ 
 $y [m]$ & -0.2 & 0.075 & 0.2\\
 $z [m]$ & 0.15 & 0.025 & 0.35 \\
 $F_{min} [N] $ & 13 & 2 & 23 \\
 \hline
\end{tabular}
\caption{}
\end{center}
\label{table:1}
\vspace{-1cm}
\end{table}

\subsection{Optometrist's Algorithm}
\label{subsection: Optometrist's Algorithm}
 
The Optometrist's algorithm (OA, see Algorithm \ref{alg:cap}) was developed and used to determine a personalized set of parameters using a technique similar to an optometrist determining the correct power of one's eye-glasses. The user was shown two options, variations of a single parameter, and was asked to choose the preferred option. Once the tuning of a parameter converged, the OA moved on to the next one. 
One of the options was the handover preferred in the previous step, and the other was a new option (with the obvious exception of the very first step). After the user chose an option, the algorithm stepped, varying the parameter by one step size ($s$). The previous choice decided the direction of the step. If the new option was chosen, the next step was taken in the same direction as the new option. 
If the same option was chosen again, then the direction for the next step was flipped. This allowed us to parse and present the entire parameter range to the user, possibly accessing parts of previously disregarded parameter space if needed. The process finished when the difference between the two options to be presented to the participant became smaller than the step size or when the same option was chosen four times in a row (this second stopping criterion is omitted in Algorithm \ref{alg:cap} for readability).
We predefined the order in which we tuned the parameters as Speed, Position $(x, y, z)$, and Force. In a few iterations of this exercise, the algorithm converged to the set of parameters that the user preferred the most.

\begin{algorithm}
\caption{The Optometrist's Algorithm}\label{alg:cap}
\begin{algorithmic}
\State $L \gets Min.$
\State $H \gets Max.$
\State $M = avg(L,H)$
\State $s \gets Step$
\State $handover(L);$ \Comment{Run handover with parameter L}
\State $handover(H);$
\State $choice \gets user\_chose(L,H)$
\If{$choice = L$}
    \State $option1 \gets L$
    \State $option2 \gets M$
\Else
    \State $option1 \gets H$
    \State $option2 \gets M$
\EndIf
\While{$|option1-option2|>s$}
\State $handover(option1);$
\State $handover(option2);$
\State $choice \gets user\_choose(option1,option2);$
\If{$choice$ is $option1$}
    \If{$option1 > option2$}
        \State $H \gets H-s$
        \State $option2 \gets H$
    \Else
        \State $L \gets L+s$
        \State $option2 \gets L$ 
    \EndIf
    \State $option1 \gets option1$
\ElsIf{$choice$ is $option2$}
    \State $temp \gets option2$
    \If{$option1 > option2$}
        \State $option2 \gets option1 -  s$
    \Else
        \State $option2 \gets option1 +  s$
    \EndIf
    \State $option1 \gets temp$
\EndIf
\EndWhile
\end{algorithmic}
\end{algorithm}

The OA presented here is optimized for small-scale problems with fewer parameters and a limited range. 
However, it is possible to tune several thousand parameters at a time, even with more extensive ranges, by introducing probabilistic steps and tuning sets of dependent parameters \cite{fusion}. In any case, setting up the parameters and their ranges takes some expert knowledge and effort.

\section{EXPERIMENT}
\subsection{Task and Protocol}
The participants performed robot-human handovers during two experiment phases, preceded by short practice sessions each. A tuning and an evaluation phase were performed sequentially in a single sitting of about 20 minutes. During the experiment, the participants were seated in a fixed chair in front of the robot, with a table being the shared workspace of the human-robot team (Fig. \ref{fig:Robot-human handover}). To obtain qualitative data, the participants were asked to complete questionnaires before and after the experiment (see Sec. \ref{section:Subjective-Metrics}). The first practice session of five handovers before the tuning phase was for participants to get used to the robot's actions and the task. The parameters used for these handovers were a fixed set of near-average parameters, the same for each participant. 

In the \textit{tuning phase}, the participants interacting with the OA were told what aspect of the handover they were tuning in layperson's terms (speed, position, and the force of the grip). The participants had to verbally announce their choice to the experimenter, who then used a simple interface to register the choice. In the second practice session of five handovers (between the tuning and the evaluation phases), the participants were asked to perform the personalized handovers. The handovers performed in these practice sessions were used to study the differences between the handovers before and after tuning. The participants were again shown two options in the \textit{evaluation phase}. One of the options was their personalized handover, and the other was a different one - similar to the preferred handover, but with slight noise added to one of the parameters. Participants were asked to identify their tuned handover. Note that as only one parameter was varied at a time and within a limited amount (1-2 times the step $s$), the handovers were still relatively similar to each other, and it was more challenging for the participants to find their own handover (compared to varying more parameters and in a broader range).

\begin{figure}[htbp]
	\centering
		\includegraphics[scale=0.4]{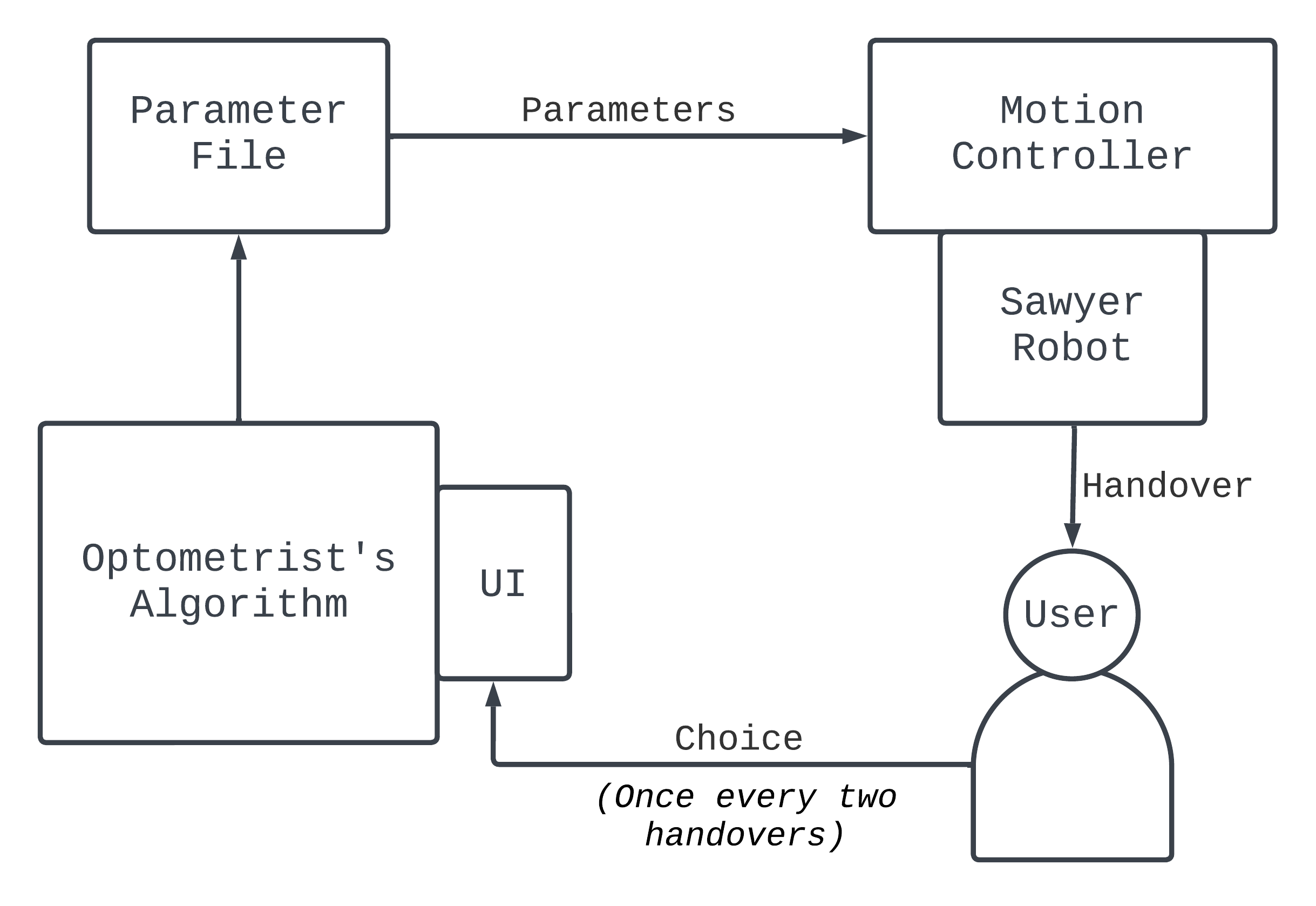}
	\caption[]{Flowchart of the 'Tuning Phase'}
	\label{fig:tuning-flowchart}
    \vspace{-0.5cm}
\end{figure}

\begin{figure*}[ht]
	\centering
		\includegraphics[scale=0.22]{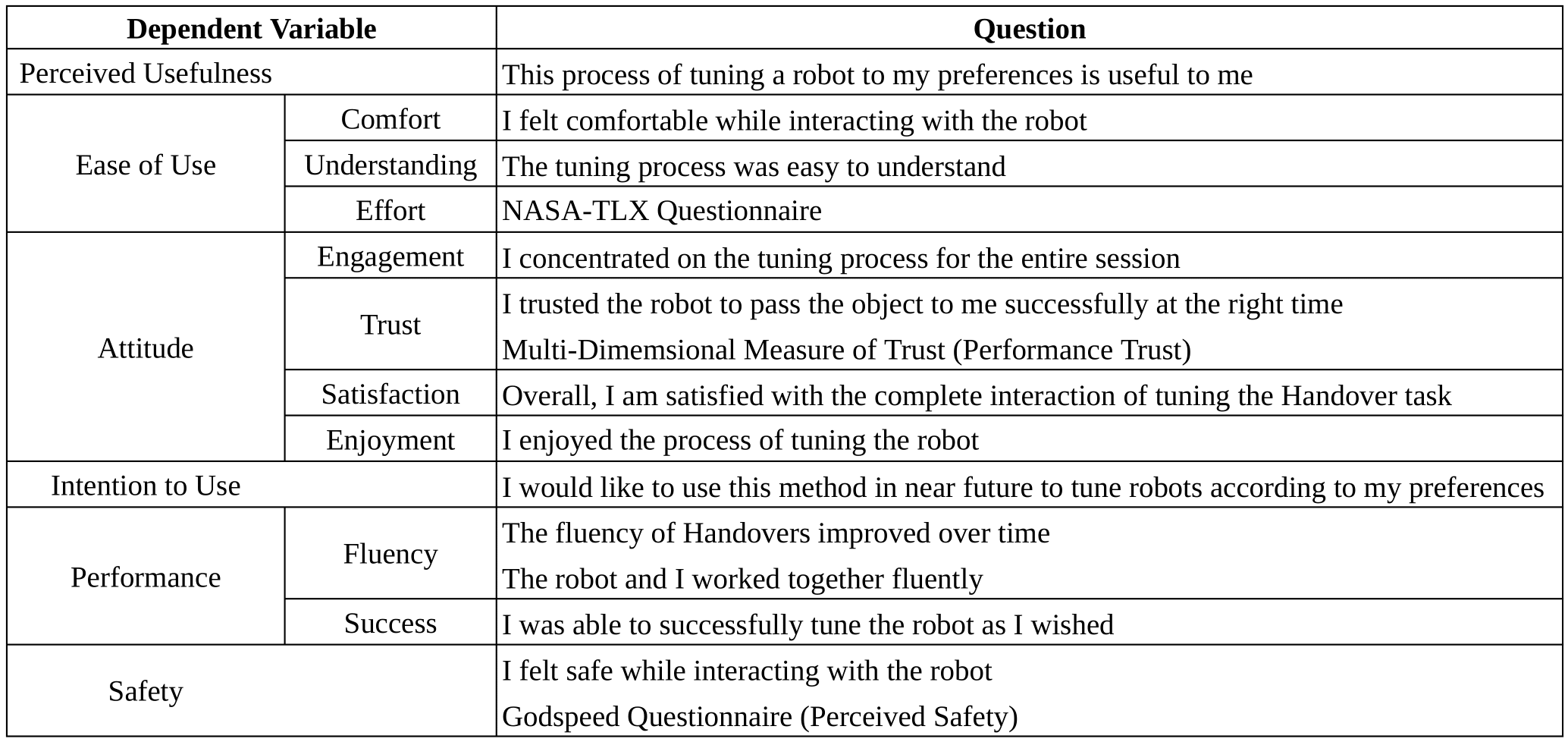}
	\caption[]{Subjective questionnaire used in this study. \cite{TAM,maya,avioz,nasa-tlx,mdmt,godspeed,hoffmann-fluency,paliga}}
	\label{fig:TAM Subjective Questionnaire}
 \vspace{-0.5cm}
\end{figure*}

\subsection{Participants}

The experiment was performed with $N=30$ participants (Ages: $19-38$, $(M)28.8 \pm(SD) 4.6$ years, $50\%$ Females), all of whom volunteered to be in the experiment for no compensation. All the participants were healthy university students or researchers. The participants were from a wide range of professional and educational backgrounds. The experiment was approved by the ethical committee of the Ben-Gurion University's Department of Industrial Engineering and Management.

\subsection{Subjective Metrics}
\label{section:Subjective-Metrics}
The participants were asked to complete the "Technology Adoption Propensity (TAP)" \cite{tap} and "Negative Attitude towards Robots Scale (NARS)" \cite{NARS} questionnaires before the experiment. After the experiment, the Technology Acceptance Model (TAM)\cite{TAM} was used to study the acceptance of the personalization task based on factors such as 'Ease of Use,' 'Perceived Usefulness,' 'Attitude' of the users, and their 'Intention of Using' this method in the future. The questionnaire used was based on questionnaires established in earlier studies\cite{maya},\cite{avioz}. The 'Effort' component of the 'Ease of Use' variable was evaluated using the NASA-TLX Questionnaire \cite{nasa-tlx} for perceived workload and effort. For 'Trust', the 'Competency' and the 'Reliability' sub-scales of the Multi-Dimensional Measure of Trust (MDMT)' \cite{mdmt} were used. All evaluations were based on a 7-point Likert scale (unless stated otherwise). 

We also studied the subjective perception of factors such as safety and fluency of the handovers. Safety was evaluated using the Godspeed Questionnaire for Perceived Safety \cite{godspeed}, and the fluency questionnaire was based on the works of Hoffmann \cite{hoffmann-fluency} and Paliga et al.\cite{paliga}. The subjective questionnaire is detailed in Fig.\ref{fig:TAM Subjective Questionnaire}

\subsection{Objective Metrics}
\subsubsection{Handover Success}
The success rate of handovers was calculated 
according to $1 - \frac{Failed\ Handovers}{Total\ Handovers}$ and included all the handovers in the tuning, practice, and evaluation phases. The success of a handover in an interactive adaptation scenario is of utmost importance, as more failures may lead to sub-par tuning. As this study included participants from varying backgrounds, with the majority having never used a robot before, the success of a handover was crucial in having the non-expert user trust the robot \cite{trust, trust-succ}. Due to this reason, the controller was designed to be robust. Hence the handover success rate was high during the experiment. It is important to note that while a high handover success rate does not imply successful use of the OA, it ensures that a failure-ridden robot interaction does not influence the users' perception of the effectiveness of the OA. Therefore we report on the general success rate throughout the experiment rather than comparing before and after the tuning phase (they are similarly high by design).
Failed handovers in the experiment were defined as any handover instances that did not result in a successful object transfer (drops, non-release of object, software, and communication failure of the robot). The experimenter manually recorded the failures during the experiment.

\subsubsection{Fluency}
Fluency is a commonly measured and studied characteristic of a human-robot interaction\cite{hoffmann-fluency}. 
In this work, we use the fluency metrics defined by Hoffmann \cite{hoffmann-fluency}. We use four metrics and correlate these metrics with the subjective perception of fluency. These metrics are defined as below:
\begin{itemize}
    \item Robot idle time (R-IDLE): Ratio of total task time spent without performing an activity by the robot.
    \item Human idle time (H-IDLE): Ratio of total task time spent without performing an activity by the human.
    \item Concurrent Activity (C-ACT): Ratio of the task time spent while both the agents perform an activity.
    \item Functional Delay (F-DEL): Ratio of the task time spent between the end of one agent's activity and before the beginning of the other agent's subsequent activity. 
\end{itemize}

While we list, define, and evaluate C-ACT, its relevance for our handover task and controller is limited. 
This is because we have programmed constant delays into the interaction, and the overlap of concurrent activity in our interaction is negligible.

\subsubsection{Needed Tuning Steps}
We counted the steps required for each parameter to be tuned using the OA. This parameter is necessary to study the user effort in tuning the controller. Each extra trial adds approximately $50$ seconds to the tuning phase and requires the user to perform two more handovers. This also indicates which parameter was easy to tune and if any parameter confused the participants.

\subsubsection{Identifying Tuned Handovers}
In the evaluation phase, we asked the users to identify their tuned handovers among similar handovers. This yielded evidence about the personalization process's success and gave us an idea of the parameters that the participants regarded highly by being able to identify minor discrepancies in them.

\section{RESULTS AND DISCUSSION}
The experiment task required an average of $18\pm1.4$ minutes and required the users to perform $58\pm4$ handovers, including practice, tuning, and evaluation phases. All users performed five handovers in each practice session and ten handovers in the evaluation session (20 out of about 60 handovers). $20\%$ participants showed a negative attitude towards robots, while $57\%$ showed a moderately positive attitude on the NARS Scale (mean score $=4.63\pm1.39$). On the TAP scale, all the participants showed a high propensity for adopting new technology (mean score $=5.18\pm1.41$).

\subsection{Success Rate}
The experiment had a very high handover success rate ranging from $94.83\%$ to $100\%$ ($mean=99.46\%$), including the handovers during tuning. 

Of the failures, three occurred because the Robot's motion controller failed to generate a collision-free trajectory. Four failures were due to false triggering of the transfer. Once, a participant dropped the object with no fault of the robot. For the failures during the tuning phase, the pair of options was repeated to continue tuning.

\subsection{Personalization}
\begin{figure*}
	\centering
    \vspace{-2cm}
		\includegraphics[scale=0.8]{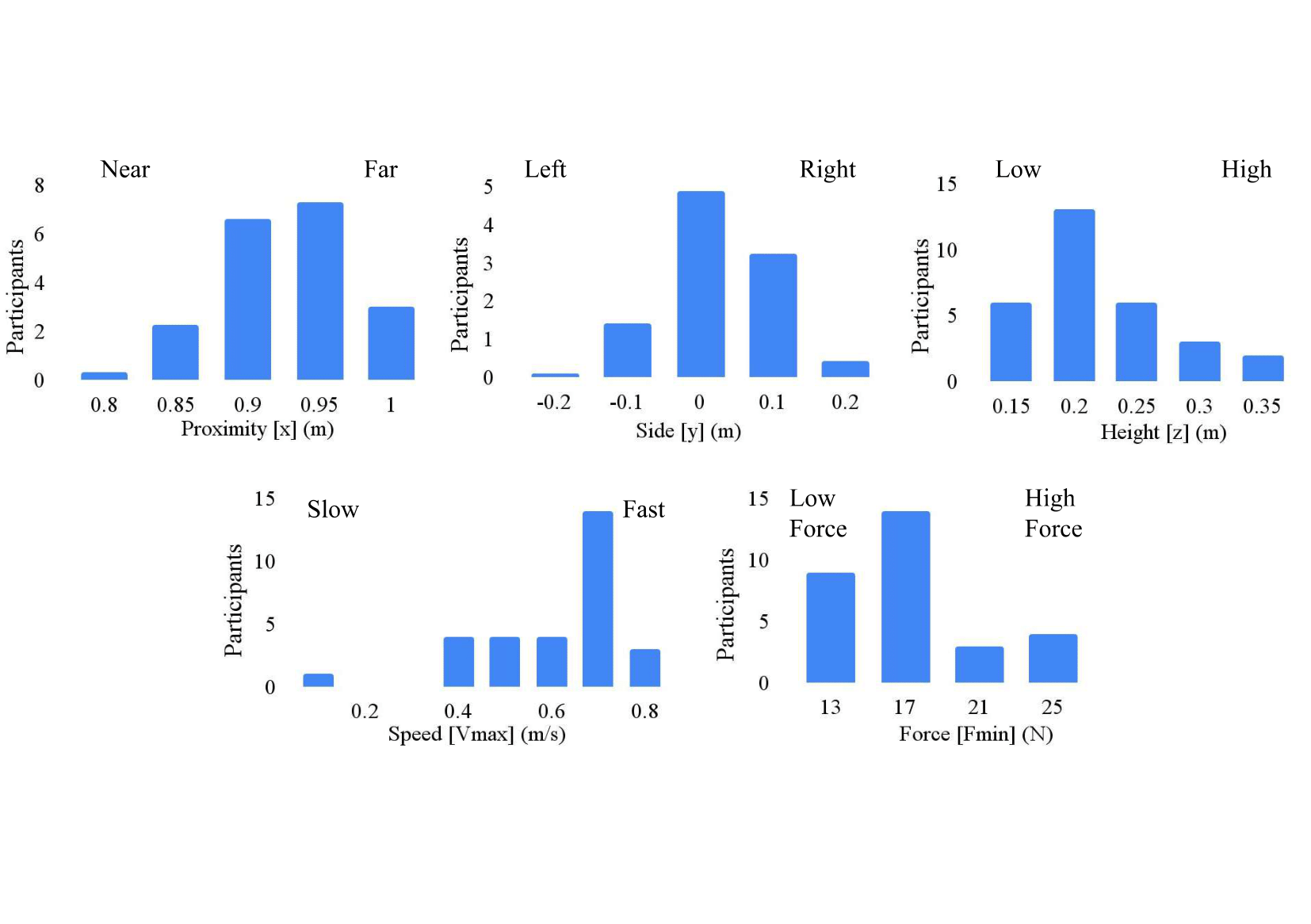}
    \vspace{-2.2cm}
	\caption[]{Histogram of all participants' parameter values after personalization -- for all tunable parameters.}
	\label{fig:charts}
\end{figure*}

The distribution of parameters preferred by the users is shown in Fig.\ref{fig:charts}. 
It is clear from the data that different participants enjoy different parameter values and can tune for them. The participants preferred to tune the handovers to be at a proximal handover location rather than the mean parameter value ($p<0.01$). But most people decided not to let the robot come too close. While choosing the side of the handover, $73.33\%$ participants preferred to keep the handover location at the center. Of the people that preferred to choose either the left or right side ($8$ out of $30$), all decided to choose the side of their dominant hand. Most participants choose to keep the handover height lower than the mean (starting) value of the parameter ($p<.001$). 
The users preferred handovers faster than the mean speed ($p<0.001$). While this parameter can also be task-dependent, even non-experienced users choose to go for higher speeds in robots. Participants preferred low values of threshold trigger force ($p<0.02$) compared to the parameter range's mean. The users tried to minimize their efforts while performing the handovers, as reported in \cite{hoffmann-fluency}. Many users admitted that they would have liked to choose a lower force than the minimum of the parameter range. However, in this study, we were limited by the design of the controller (robustness). 

The entire tuning process took between $10-12$ minutes and $18$ and $21$ steps ($mean=19.2\pm1.5$), amounting to twice the number of handovers performed. Tuning the handover height required the most steps ($4.1\pm0.9$ and $143\pm32$ sec) while tuning the side of the handover location (left or right) required the least steps ($3.2\pm0.5$ steps and $110\pm21$ sec). Despite every user needing to perform approximately $40$ handovers for the personalization, the users perceived this activity as a low workload on the NASA-TLX Questionnaire (score$=2.68\pm1.49$ on a 7-point Likert scale). 

\subsection{Evaluation Phase}
Most participants could identify their handover between variations in speed ($mean\ score = 0.75,$ out of $1$), followed by variations in threshold force values required to trigger the transfer ($mean\ score=0.58$ out of $1$). For variations in handover location, the participants performed the worst ($mean\ score = 0.53$ out of $1$).
In five attempts, the average participant scored $3.19\pm0.92$. The participants who performed the best were those who had tuned the values to the extreme ends of the parameter range.

\subsection{Subjective Measures}
\begin{figure}[htbp]
	\centering		\includegraphics[scale=0.65]{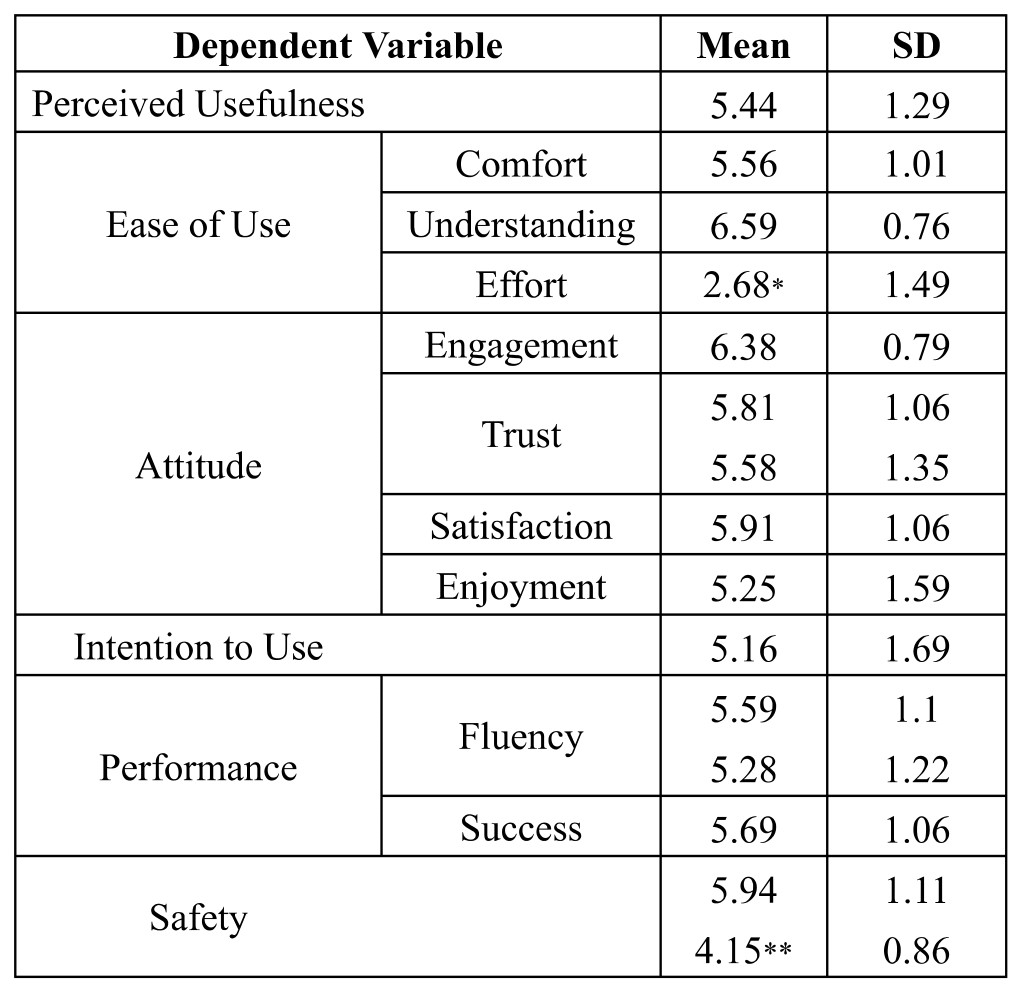}
	\caption[]{Subjective Questionnaire Scores \footnotesize{($^{*}$inverse scale, $^{**}$five-point scale)}}
    	\label{fig:TAM Subjective Questionnaire scores}
\vspace{-.5 cm}
\end{figure}

The users evaluate the personalization process very positively through the subjective questionnaire (Fig.\ref{fig:TAM Subjective Questionnaire scores}). The process is primarily perceived as useful ($5.44\pm1.29$) with $19$ out of $30$ users. $21$ out of $30$ users agreed that the method was easy to understand. The users also showed a positive attitude towards this method, especially in its 'Engagement' component, with $15$ out of $30$ participants strongly agreeing to have been completely focused on the activity. Users ($87\%$, $26$ out of $30$) generally agreed ($5.16\pm1.69$) that they might use this method in the future.
Of the four users who disagreed, one reported that the tuning process did not produce a desirable handover. Two users mentioned that the process was \textit{"too repetitive"} and was \textit{"not enjoyable"}, while one user felt that \textit{"the robot did not respond to their feedback"}.

On the Multi-Dimensional Measure of Trust (MDMT) scale, the user scored the robot highly on both the Reliability ($5.66 \pm 1.41$) and the Competency sub-scales ($5.49\pm1.29 $). The users also perceived high robot safety, scoring $4.14\pm0.86$ on a five-point scale of the Godspeed questionnaire. The user perception of trust and safety did not change even for those participants who experienced one of the few handover failures. Few users reported that they were at times surprised by some of the options that were \textit{"too fast"} or when the robot \textit{"came too close"}. However, all the users rated the task highly on the trust and safety scales. A low correlation was observed between success rate and safety ($r=0.20$) and between success rate and trust ($r=0.15$). 

\subsection{Fluency}
Our metrics have suggested a high perception of both subjective fluency ($5.28\pm1.22$) and improvement in fluency ($5.59\pm1.10$) after tuning. The significance of all the metrics was evaluated using a two-tailed test. A significant increase in H-IDLE ($p<0.01$) and a highly significant decrease in F-DEL $(p<0.01)$ was observed after personalization. Both these observations correspond to an increased objective fluency after personalization \cite{hoffmann-fluency}. Comparing fluency metrics to subjectively perceived fluency, there was no correlation between either change in H-IDLE ($r=0.096$) or F-DEL ($r = -0.092$). This implies that the handover fluency improved objectively, but not all participants could perceive it. An inconsistent subjective perception of fluency in a proactive coordination task, such as ours (the robot did not wait for a signal from the human before starting the handover), is already reported by Huang et al.\cite{huang}. 

There was no significant change in C-ACT ($z=1.17, p>0.1$), which was expected, as the delays programmed in the controller were constant, to make the tuning interaction more consistent. After personalization, R-IDLE showed a highly significant decrease ($p<0.01$). A major factor contributing to R-IDLE was the robot waiting for the user to initiate the object transfer. Initially, as the handover location was not personalized and the force to trigger the transfer was not preferred by most users, this phase took longer. This metric can also be interpreted as task efficiency in terms of robot usage during the task. This task efficiency improves significantly after personalization.

\section{CONCLUSIONS}

In this work, we presented using the Optometrist's Algorithm to personalize robot-human handovers within a few interactions. We showed that our algorithm provides non-expert users with an easy, engaging, effortless, and, thus, effective method to personalize their handovers intuitively. We successfully tuned intuitive parameters like speed and non-intuitive parameters like the handover pose. Personalization appears successful as participants cannot only tune their controllers to their liking but also differentiate between their control settings and perturbed versions of their controllers. On a more fundamental level, all our participants end up with individual controllers and utilize various parameters. This is again in contrast to the most optimal solution according to objective metrics, where the robot moves, e.g., as fast as possible to the participants. 

We saw that most participants preferred a quick and effortless handover that was proximal but not too close to them. The handover speed was the most important parameter affecting user preference. Our human-in-the-loop personalization process also improved the objective fluency of the handovers. This did not fully reflect in the subjective metrics -- which we attribute to the nature of our task (being proactive). Also, user preferences can generally depend on the specific task or be object-specific. Both are possible variations of our study that remain for future work.

A further variation would be to apply the algorithm to tune several parameters at once, which is especially useful for a larger number of parameters. While the potential of this class of algorithms to tune a large number of parameters has already been established in other fields\cite{fusion}, its application and effectiveness for tuning a robot's behavior, with a non-expert user in the loop, when dependent on many more and larger parameter spaces then we used, remains to be shown.

\section*{ACKNOWLEDGMENT}
This research was supported by Ben-Gurion University of the Negev through the Agricultural, Biological, and Cognitive Robotics Initiative, the Marcus Endowment Fund, the W. Gunther Plaut Chair in Manufacturing Engineering, ISF Grant 1651/19, and the Lynn and William Frankel Center for Computer Science, the Israeli Ministry of Aliyah and Integration as well as BITS-Pilani K. K. Birla Goa Campus, India.


\end{document}